\documentclass{article}
\usepackage{ijcai21}

\usepackage{times}

\usepackage{soul}
\usepackage{url}
\usepackage[hidelinks]{hyperref}
\usepackage[utf8]{inputenc}
\usepackage[small]{caption}
\usepackage{graphicx}
\usepackage{amsmath}
\usepackage{booktabs}
\usepackage{subcaption}
\usepackage{color,xcolor}
\usepackage{algorithmic}
\usepackage[ruled]{algorithm2e}
\usepackage{amssymb}
\usepackage{setspace}

\title{Self-Regression Learning for Blind Hyperspectral Image Fusion Without Label}

\author{Wu Wang, Yue Huang, Xinhao Ding}

\begin{document}

\maketitle

\begin{abstract}
Hyperspectral image fusion (HIF) is critical to a wide range of applications in remote sensing and many computer vision applications. Most traditional HIF methods assume that the observation model is predefined or known. However, in real applications, the observation model involved are often complicated and unknown,  which leads to the serious performance drop of many advanced HIF methods. Also, deep learning methods can achieve outstanding performance, but they generally require a large number of image pairs for model training, which are difficult to obtain in realistic scenarios. Towards these issues, we proposed a self-regression learning method that alternatively reconstructs hyperspectral image (HSI) and estimate the observation model. In particular, we adopt an invertible neural network (INN) for restoring the HSI, and two fully-connected network (FCN) for estimating the observation model. Moreover, \emph{SoftMax} nonlinearity is applied to the FCN for satisfying the non-negative, sparsity and equality constraints. Besides, we proposed a local consistency loss function to constrain the observation model by exploring domain specific knowledge. Finally, we proposed an angular loss function to improve spectral reconstruction accuracy. Extensive experiments on both synthetic and real-world dataset show that our model can outperform the state-of-the-art methods.
\end{abstract}

\section{Introduction}

Hyperspectral image (HSI) contains rich spectral and spatial information, which have wide range of applications in both computer vision and remote sensing communities. However, due to physical limitation, HSI usually has low spatial resolution compared with multispectral cameras, which hampers its practical usage. The purpose of HIF is to reconstruct a HR-HSI ($X\in {\mathbb{R}}^{(W\times H\times C)} $) with a pair of LR-HSI ($Y\in {\mathbb{R}}^{(w\times h\times C)}$) and corresponding HR-MSI ($Z\in {\mathbb{R}}^{(W\times H\times c)}  $), where $W$, $H$ and $C$ denotes the image width, image height and number of spectral bands, respectively. As $w\ll{W}, h\ll{H}, c\ll{C}$, HIF is a highly ill-posed problem, thus difficult to solve. Previous conventional works generally depend on the knowledge of observation model, such as spectral response function (SRF) and point spread function (PSF) of sensors. The observation model of HIF is usually formulated by following linear functions:
\begin{align}
\mathbf{Y} &= \mathbf{X}\mathbf{B}\mathbf{S}, \label{equ:e1}
\\
\mathbf{Z} &= \mathbf{R}\mathbf{X},\label{equ:e2}
\end{align}
However, in practical applications, the information about sensors is often insufficient or unknown. Therefore, the blind HIF is the key problem to be solved in the current field.

HIF methods based on deep learning can directly learn the required mapping without the knowledge of observation model. However, the demand for sufficient training image pairs inevitably limits their practicability. In addition, the general deep neural network has the problem of information loss, which makes the preservation of spatial and spectral information more difficult. To address these challenges mentioned before, we propose a novel self-regression network (SRFN) for blind HIF that achieve SOTA performance.

Our contributions can be summarized as follows:
\begin{itemize}
	\item We proposed a self-regression network that alternatively estimating both the observation model and fusion processes, which enable blind HIF without labeled data for training.
	\item To estimating the fusion process, we proposed an invertible neural network (INN) to reduce the inherently information loss of general deep neural network for better preserving the spatial and spectral information.
	\item In order to constrain the observation model, we introduce a local consistency loss function by considering domain specific knowledge. An angular loss is also introduced to improve the spectral reconstructing accuracy.
	\item Extensive experiments on synthetic and real-world datasets show that our model can achieve state-of-the-art performance compared with the latest unsupervised blind HIF methods.
\end{itemize}

\section{Related Works}

\subsection{Conventional Methods}
In recent years, there are many conventional methods for HIF. Inspired by task of spectral unmixing, several matrix factorization approaches~\cite{Alpher014,Alpher015,Alpher016,Alpher017,Alpher021} are proposed. Since the matrix data representation is difficult to fully exploit the inherent HSI spatial-spectral structures, several approaches~\cite{Alpher018,Alpher019} based on tensor factorization are designed to better model the spatial structure of HSI. Besides, a set of penalty based approaches~\cite{Alpher011,Alpher012,Alpher013,Alpher031,Alpher091} are also brought out by exploiting different prior knowledge.

Most of these conventional methods are non-blind methods. As far as we know, HySure~\cite{Alpher012} is the only conventional method that able to learn the observation model from unlabeled data. HySure attempts to estimates the observation models via convex optimization based on two quadratic data-fitting terms and total variation regularization. However the total variation regularization may bring over-smooth effect.

\subsection{Deep Learning Methods} In the past few years, various HIF technologies based on CNN have been proposed. Most of these methods require either large training data~\cite{dbin} or the knowledge of observation model~\cite{dhsis}, which are both unrealistic in real HIF scenario. Very recently, ~\cite{zhang} and~\cite{cucanet} were proposed to adaptively learn the two functions of observation model for unsupervised and blind HIF. However, neither of them pay attention to the spectral and spatial information loss cased by general neural networks, which is undesirable for HIF. In addition, in order to solve the highly unconstrained problem, they minimize multiple domain specific loss functions, which makes them difficult to select the optimal hyperparameters.

\section{Method}
\subsection{Motivation} 
\begin{align}
\mathbf{Y} &= \mathbf{X}\mathbf{B}\mathbf{S}, \label{equ:e1}
\\
\mathbf{Z} &= \mathbf{R}\mathbf{X},\label{equ:e2}
\end{align}
where $B$ stands for a convolutional operation between the point spread function of the sensor and the HR-HSI bands, $S$ signifies a downsampling operation, and $R$ is the spectral response function of the multi-spectral imaging sensor. Most of the conventional works make assumptions that $B$ and $R$ are known beforehand. Often they tries to solve this problem via minimizing:
\begin{equation}\label{equ:e3}
\mathop{\min}_{\mathbf{X}}~ {\left\|\mathbf{Y} - \mathbf{XBS}\right\|^2_F} + \mathbf{\lambda_1}{\left\|\mathbf{Z} - \mathbf{RX}\right\|^2_F} + \mathbf{\lambda_2\phi(X)},
\end{equation}
where the first and second terms enforce data fidelity, and the third term is a regularization term. $\lambda_1$ and $\lambda_2$ are regularization parameters, and $F$ represents for Frobenius norm. However the information about the sensors maybe unavailable or inaccurate, that is the $B$ and $R$ maybe unknown, so the practicality of these methods could be limited.

Another straightforward strategy for blind HIF is to directly learn a mapping function between of the two inputs $[Y,Z]$ and output $X$ via off the deep neural networks such as ResNet\cite{Alpher044}. This can be done by optimizing a simple objective function:

\begin{equation}\label{equ:l1}
\mathop{\min}_{\mathbf{\theta}}~ {\left\|\mathbf{X} - \mathbf{f}_\theta{\mathbf{(Y,Z)}}\right\|^2_F},
\end{equation}
where $\theta$ represents the weights of CNN. However, most deep learning based method require a large number of labeled data for training, which is hard to obtain for hyperspectral imaging. 

To solve these problems, we proposed a deep neural framework to both learn the fusion process and observation model. The fusion process and observation models can be written as 
\begin{align}
\mathbf{\hat{X}} &= \mathbf{f(Y,Z)},
\label{equ:fusion}
\\
\mathbf{\hat{Y}} &= \mathbf{g_B(\hat{X})},
\label{equ:ober_B}
\\
\mathbf{\hat{\textbf{Z}}} &= \mathbf{g_R(\hat{X})},
\label{equ:observation}
\end{align}
where $f$ means the fusion process, and $g$ represents the observation model. A simple way to simultaneously learn $f$ and $g$ is to minimize the following object function:
\begin{small}
\begin{equation}\label{equ:srfn}
\mathop{\min}_{\mathbf{\hat{X}}}~ \mathcal{L}(\mathbf{Y}, \mathbf{g_B(f{(Y,Z))}}) + \mathcal{L}(\mathbf{Z}, \mathbf{g_R(f(Y,Z))}) +\mathbf{\lambda_{B}\phi_{B}(B)}+\mathbf{\lambda_{R}\phi_{R}(R)},
\end{equation}
\end{small}
where the first and second terms enforce data fidelity, the third and fourth terms impose prior knowledge on the observation model. $\mathcal{L}$ represents the reconstruction loss function, $\lambda_B$ and $\lambda_R$ are regularization parameters. Minimizing this object function form a self-regression pattern, thus our method is called self-regression learning. Inspired by ~\cite{dbin}, we further proposed a three stage self-regression learning fashion, as shown in Algorithm.~\ref{algorithm1}.

\begin{algorithm}[t]
	\caption{Three stage self-regression learning.} \label{algorithm1}
	\KwIn{$Y$ (LR-HSI); $Z$ (HR-MSI)}
	\For{$n=1$ to $3$}
	{
		\If{$n=1$}
		{Compute $\mathbf{\hat{X}}^1 = \mathbf{f_1(Y,Z)}$\\
		 Compute $\mathbf{\hat{Y}}^1  = \mathbf{g_B(\hat{X}^1)}, \mathbf{\hat{Z}}^1  = \mathbf{g_R(\hat{X}^1)}$}
		\If{$n=2$}
		{Compute $\Delta\mathbf{\hat{X}}^2 = \mathbf{f_2(Y-\hat{Y}^1,Z-\hat{Z}^1)}$\\
			Compute
			$\mathbf{\hat{X}}^2 = \mathbf{\hat{X}}^1+\mathbf{\Delta{\hat{X}}^2}$\\
			Compute $\mathbf{\hat{Y}}^2  = \mathbf{g_B(\hat{X}^2)}, \mathbf{\hat{Z}}^2  = \mathbf{g_R(\hat{X}^2)}$}
		\If{$n=3$}
		{Compute $\Delta\mathbf{\hat{X}}^3 = \mathbf{f_3(Y)-\hat{Y}^2,Z-\hat{Z}^2)}$\\
			Compute
			$\mathbf{\hat{X}} = \mathbf{\hat{X}^2+\Delta{\hat{X}}^3}$\\
			Compute $\mathbf{\hat{Y}}  = \mathbf{g_B(\hat{X}), \hat{Z}}  = \mathbf{g_R(\hat{X})}$}
	}
	\KwOut{$\hat{X}$ (HR-HSI), $\hat{Y}$ (LR-HSI), $\hat{Z}$ (HR-MSI)}
\end{algorithm}

In stage 1, we obtain an initial HR-HSI. Then the initial HR-HSI is used to reconstruct a coarse LR-HSI and HR-MSI. Since the reconstruct results may be not accurate, so we further refine them in stage 2 and stage 3 with the reconstruction residual. We argue that compared to the single stage counter partner, the proposed three stage modification enables us to apply deeper neural network for learning fusion process, and is much easier to train. 

\begin{figure*}[!h]
	\centering
	\includegraphics[width=.90\textwidth]{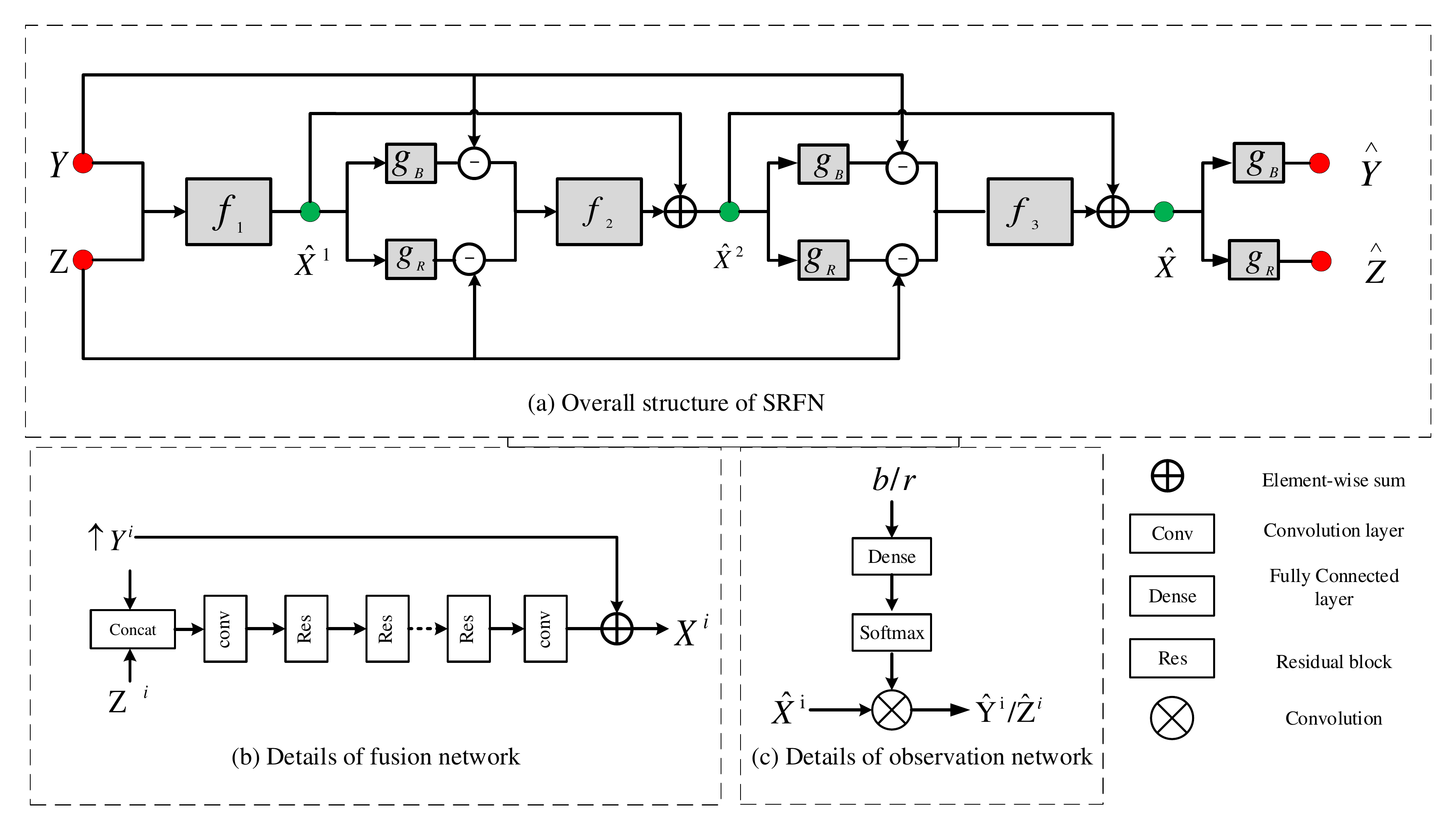}
	\caption{The detail structure of SRFN. (a) (top) Overall structure of SRFN, $f_i$ represents the fusion network, number $i$ denotes the stage. $g_B$ and $g_R$ are observation networks that are used to learn $B$ and $R$ respectively. (b) (bottom-left) The detail structure of fusion network. (c) (bottom-right) The detail structure of observation networks.}
	\label{fig:SRFN}
\end{figure*}

\subsection{Network Architecture} The detailed construction of the proposed network is displayed in Fig.~\ref{fig:SRFN}. The core insight of SRFN are the fusion network and the observation network.

\paragraph{Fusion Network with invertible ResBlock} The fusion networks are response for reconstructing and refining HR-HSI. Since the three fusion networks share the same structure, we only demonstrate the detail structure of of one network. The detail network construction is depicted in right part of Fig.~\ref{fig:SRFN}(b). As the spatial resolution of MR-HSI is much higher than LR-HSI, we first concatenate the upsampled residual of the LR-HSI with the residual of the HR-MSI. A convolution layer is further applied to match the feature dimension. Then we apply several resblocks~\cite{Alpher044} to extract features. Finally, we use a convolution layer for transforming the feature to image domain. General deep neural networks have difficulty in recovering images from their hidden representations due to the information loss\cite{Visualizing}. This is undesirable as the main goal of HIF is to preserve spectral and spatial information. The INNs are capable of recovering input images from the features, thus avoid information loss. Inspired by INNs, to counteract the inherent information loss, we choose to make resblock invertible. The work of ~\cite{invertiblenet} shown that simply adding a spectral normalization to the weights of resblocks could make it invertible. This can be formulated as:
\begin{equation}\label{equ:sn}
\mathbf{\hat{W}_i} = \mathbf{\lambda{\times{SN}(W_i)}}, \text{ and} \ 0<\lambda<1
\end{equation}
where $W_i$ denotes the weight of residual block, \emph{SN} represents the spectral normalization~\cite{SN} which is widely used for stabilizing GAN~\cite{gan} training, $\lambda$ represents a small coefficient, which is set to 1 in standard spectral normalization. To make resblock invertible, we set $0<\lambda<1$ to constrain the lipschitz constant of the resblock below one. Although spectral normalization is only an approximate estimate of lipschitz constant, setting $0<\lambda<1$ does not guarantee the invertibility of the resblock. We will show that this method does bring benefit.
\paragraph{Observation networks} The observation networks is response for learning $B$ and $R$. Matrix $B$ is a spatial blurring kernel representing the hyperspectral sensor’s point spread function, and $R$ is the  spectral response function. Usually, both $B$ and $R$ are assumed to be sparse, non-negative, and would sum up to 1. In addition, $B$ is supposed to be band independent. We adopt a pair of fully-connected networks (FCN) to serve as $g_B$ and $g_R$, respectively. As shown in Fig.~\ref{fig:SRFN} (c), the $g_B$ and $g_R$ shares the similar structure. The network firstly takes a random initialized variable $b$ or $r$ as input. After the fully connected layer, the \emph{SoftMax} non-linearity is applied to the output layer to guarantee the non-negative and sum up to 1 constraints. Finally, the learned $B$ and $R$ is applied to the generated HSI for reconstructing LR-HSI or HR-MSI via convolution operator. We set the spatial blurring kernel as $14\times14$ for all the experiments. While previous methods usually design several loss functions to guarantee these constraints, which makes picking the right hyperparameters troublesome. We show that simply introduce the \emph{SoftMax} nonlinearity is well enough to guarantee these constraints.
\subsection{Loss function} Since we do not have access to the ground truth HR-HSI, we choose to minimize the following $\mathcal{L}1$ loss: 
\begin{equation}\label{equ:re_spa_loss}
\mathcal{L}_{spa}= \|\mathbf{\hat{Y}} - \mathbf{Y}\|_1 + \|\mathbf{\hat{Z}} - \mathbf{Z}\|_1,
\end{equation}
The $\mathcal{L}1$ loss is only a measurement of spatial reconstruction accuracy. Since the HSI contain very rich of spectral information, the spectral reconstruction accuracy is equally important to spatial reconstruction accuracy. To obtain better spectral reconstruction accuracy, we apply the angular loss (SAM) to LR-HSI. The angular loss can be defined as:
\begin{equation}\label{equ:re_spe_loss}
\mathbf{\mathcal{L}_{spe}} = \mathbf{\frac{1}{MN}\sum_{i=1}^{MN}\arccos(\frac{Y^i\hat{Y^i}}{\|Y^i\hat{Y^i}\|_2})}
\end{equation}
where $M,N$ are the width and height of the LR-HSI.
Recall the observation model $Y=XBS$, multiply both sides of the equation by the matrix $R$, and substitution $Z=RX$, we have:
\begin{equation}\label{equ:local_consistency}
\mathbf{RY} = \mathbf{ZBS}
\end{equation}

Inspired by this equation, We further proposed a local consistency loss to constrain the observation model:
\begin{equation}\label{equ:local_consistency_loss}
\mathbf{\mathcal{L}_{lc}} = \|\mathbf{RY} - \mathbf{ZBS}\|_1
\end{equation}

By combining all the above-mentioned loss terms, the final objective functions of the proposed SRFN can then be expressed as
\begin{equation}
\mathbf{\mathcal{L}} = \mathbf{{\mathcal{L}_{spa}} + \beta\mathcal{L}_{spe} + \gamma\mathcal{L}_{lc}}
\end{equation}
where $\beta$ and $\gamma$ are parameters that balance the trade-off between each terms.

\section{Experiments}

In this section we present the experimental results of demonstrated methods described above.
\subsection{Setup}
\subsubsection{Datasets and Metrics}
Three publicly available hyperspectral databases were used for simulation experiments: CAVE~\cite{Alpher026}, Pavia, and WV2.\footnote{\url{https://www.harrisgeospatial.com/Data-Imagery/Satellite-Imagery/High-Resolution/WorldView-2}}

The Pavia was acquired by ROSIS airborne sensor over the University of Pavia, Italy, in 2003. The original HSI comprises $610\times340$ pixels and 115 spectral bands. We use the top-left corner of the HSI with $336\times336$ pixels and 93 bands, covering the spectral range from 380 nm to 840 nm. The CAVE contains 32 indoor images recorded under controlled illumination. Both of the CAVE and Pavia contain only HR-HSI images, while the WV2 contains a pair of real multi-spectral image and RGB image. So We use CAVE and Pavia for simulation experiments, and WV2 for real experiments.

The HR-MSI of CAVE were simulated by integrating over the original spectral channels using the spectral response of a Nikon D700 camera \footnote{\url{http://www.maxmax.com/spectral_response.htm}}. We use the code of HySure to generate the HR-MSI for Pavia \footnote{\url{https://github.com/alfaiate/HySure}}. We consider scale factors 8 for downsampling to simulate the LR-HSI for CAVE and Pavia dataset.

PSNR, SSIM~\cite{Alpher028}, SAM~\cite{Alpher029} and ERGAS~\cite{Alpher031} are used for quantitative evaluation. SAM measures the spectral difference between the estimated image and the ground truth, SSIM is an indicator for spatial structures consistency, and ERGAS reflects the overall quality.

\subsubsection{Training Setups}
Following previous works, we use isotropic Gaussian blur kernels to simulate the PSF. The range of kernel width is fixed to 1, and the kernel size is fixed to $8\times8$. For CAVE and Pavia, we directly take the whole image for training. As for the WV2, the LR-HSI are cropped to $128\times128$ image patches for training, and no data augmentations were used. The parameter $\beta$ and $\gamma$ are empirically set to 0.01 and 30, respectively. Adam optimizer is used with default setting. We train SRFN for $55000$ iterations on CAVE, and the learning rate is initialized to $2\times10^{-4}$. The SRFN is trained for $20000$ iteration on WV2 and $10000$ iteration on Pavias, while the initial learning rate is set to $4\times10^{-4}$. Through all the experiments, the training batchsize is set to 1. All the models are trained on a server equipped with NVIDIA RTX 2080 Ti GPU.

\subsection{Ablation study}
\subsubsection{Comparison of SRFN Configurations}
Table~\ref{tab:ablation} summarizes the quantitative comparisons of different SRFN configurations. The baseline is a network with the same structure as SRFN but without spectral-normalization and \emph{SoftMax} layer, and trained only with the spatial reconstruction loss. In these configurations, S represents the observation network containing \emph{SoftMax} layer (as shown in Fig.~\ref{fig:SRFN}(c)), and N marks spectral-normalization with a small coefficient. L and A denotes the network trained with the local consistency loss and spectral reconstruction loss, respectively. As in Table~\ref{tab:ablation}, all the proposed methods improve the performance over the baseline configuration. Among the proposed method, the spectral-normalization with a small coefficient achieves largest performance gain.

\begin{table}[!h]
	\centering
	\begin{tabular}{lllll}
		\hline
		Method & PSNR$\uparrow$ & SSIM$\uparrow$ & SAM$\downarrow$ & ERGAS$\downarrow$ \\
		\hline
		BaseLine  & 24.50  & 0.558 & 0.558 & 7.789 \\
		SRFN$\_$S & 29.95  & 0.818 & 5.145 & 2.512 \\
		SRFN$\_$SN & 39.59 & 0.976 & 2.830 & 0.856\\
		SRFN$\_$SNL & 40.71 & 0.979  & 2.690 & 0.861 \\
		SRFN$\_$SNLA & {\color{red}{41.67}} & {\color{red}{0.984}} & {\color{red}{2.30}} & {\color{red}{0.820}} \\
		\hline
	\end{tabular}
	\caption{Average performance for various SRFN configurations on Pavia dataset. The best results are highlighted in red color.}
	\label{tab:ablation}
\end{table}

\subsubsection{Analysis Spectral Normalization Coefficiency}
\begin{figure}[!h]
	\centering
	\includegraphics[width=.44\textwidth]{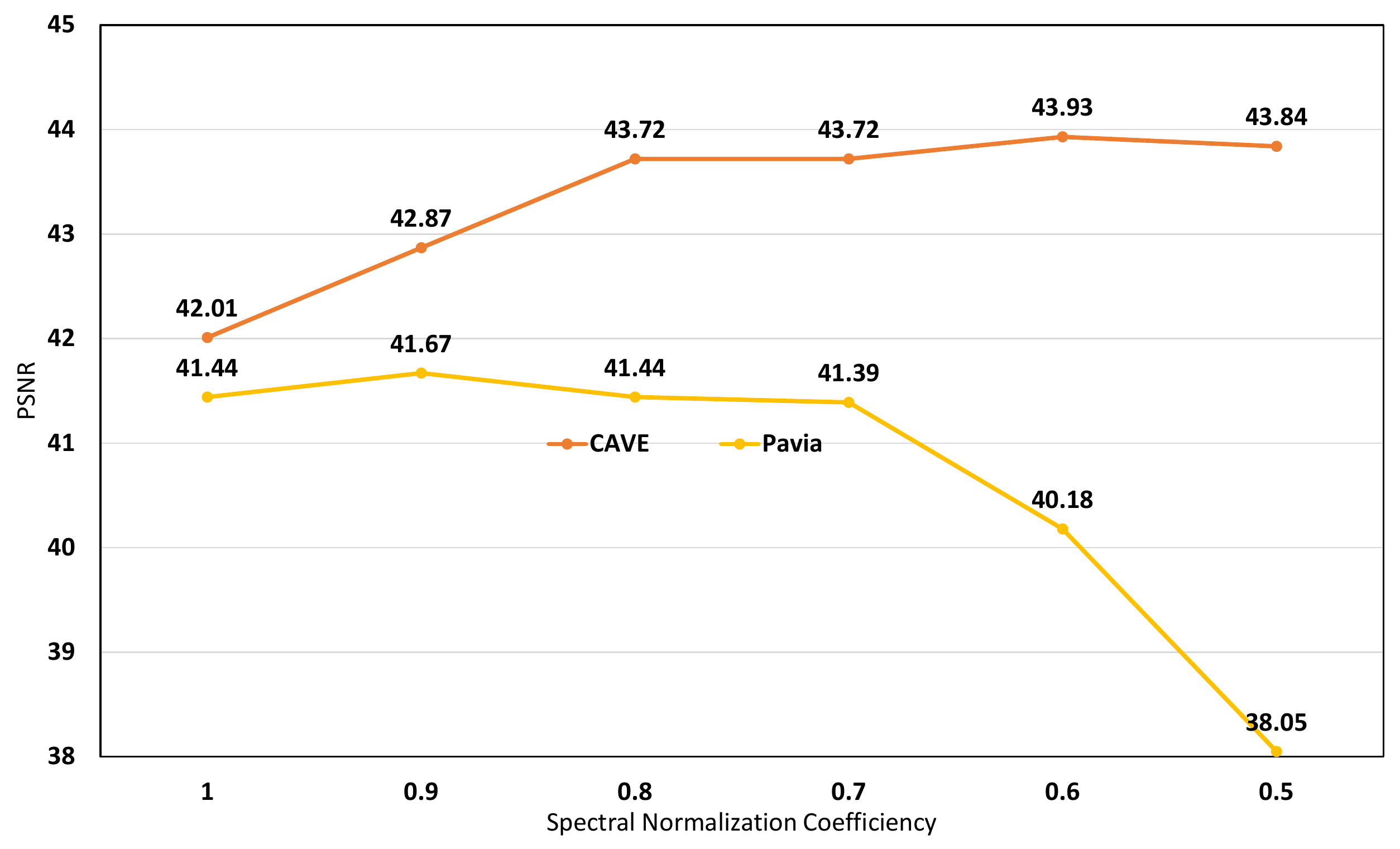}
	\caption{Comparison of SRFN of same architecture with various coefficients $\lambda$ on CAVE and Pavia dataset}
	\label{fig:sn}
\end{figure}

Decreasing the coefficient will make the lipschitz constant smaller, thus increases the ability of information preserving. Simultaneously, constraining the constant will also reduce the richness of extracted features thus decreasing the representability of network. This imposes a trade-off between information preserving and information extracting. To perform detail analysis on the spectral normalization coefficient of $\lambda$, we conduct experiments with different coefficient $\lambda$ without changing the network architecture. The experiment results are shown in Fig~\ref{fig:sn}, from which we can draw some conclusions. Firstly, applying standard spectral normalization ($\lambda=1$) already brought great performance gains. Secondly, modifying $\lambda$ will change the model performance obviously. In detail, when $\lambda$ is reduced gradually, the performance is improved first and then decreased. Finally, the optimal value of $\lambda$ is different for different datasets. Experiment results verified the trade-off between information preserving and information extracting. In the following experiments, we set $\lambda=0.6$ for CAVE, $\lambda=0.9$ for Pavia, and $\lambda=0.7$ for WV2.

\subsection{Experiments on Synthetic Indoor Images}
We first conduct experiments on 12 indoor images of the CAVE dataset. We compare our model not only with unsupervised blind HIF method, namely HySure~\cite{Alpher012} and CuCaNet~\cite{cucanet}, but also with unsupervised non-blind conventional method NSSR~\cite{nssr}. CuCaNet (ECCV 2020) is the SOTA deep learning method of blind HIF that training without label. Given the ground truth observation model, the proposed SRFN is also able to perform non-blind HIF, we denote this non-blind version as \emph{SRFN/GT}. 

\begin{figure*}[!t]
	\centering
	\begin{minipage}[t]{0.12\linewidth}
		\centering
		\includegraphics[width=0.99999\columnwidth]{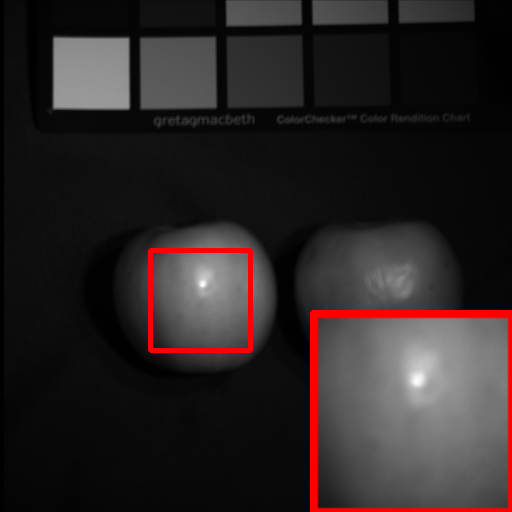} \\
		\quad
		\includegraphics[width=0.99999\columnwidth]{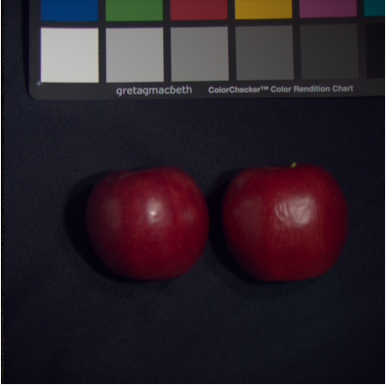}
		\subcaption{GT and RGB}
	\end{minipage}
	\begin{minipage}[t]{0.12\linewidth} 
		\centering
		\includegraphics[width=0.99999\columnwidth]{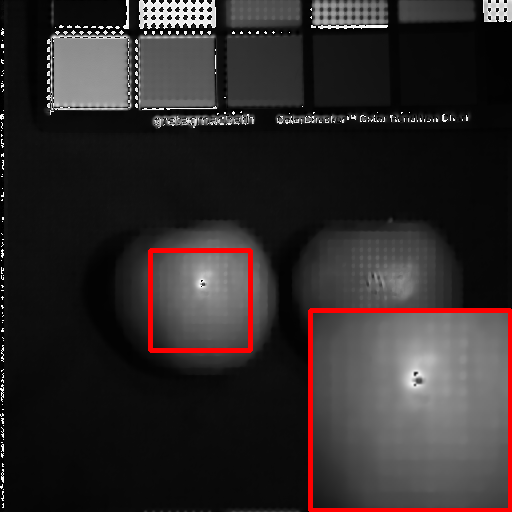} \\
		\quad
		\includegraphics[width=0.99999\columnwidth]{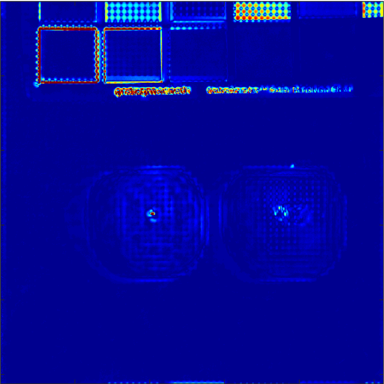}
		\subcaption{HySure}
	\end{minipage}
	\begin{minipage}[t]{0.12\linewidth}
		\centering
		\includegraphics[width=0.99999\columnwidth]{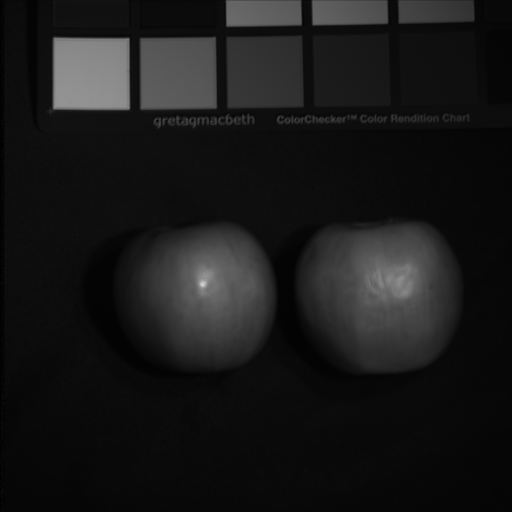} \\
		\quad
		\includegraphics[width=0.99999\columnwidth]{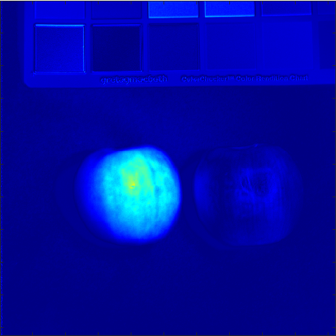}
		\subcaption{CuCaNet}
	\end{minipage}
	\begin{minipage}[t]{0.12\linewidth}
		\centering
		\includegraphics[width=0.99999\columnwidth]{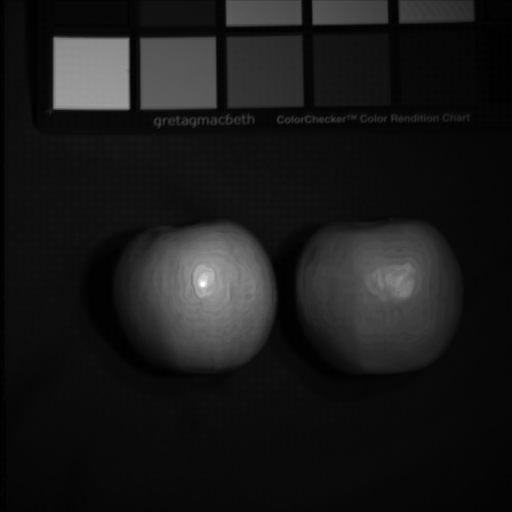} \\
		\quad
		\includegraphics[width=0.99999\columnwidth]{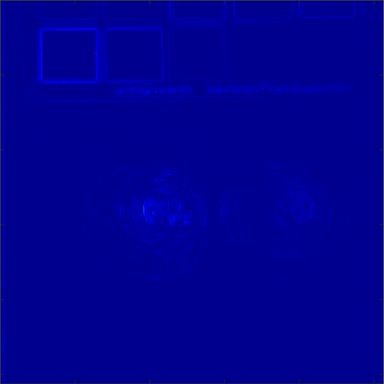}
		\subcaption{NSSR}
	\end{minipage}
	\begin{minipage}[t]{0.12\linewidth}
		\centering
		\includegraphics[width=0.99999\columnwidth]{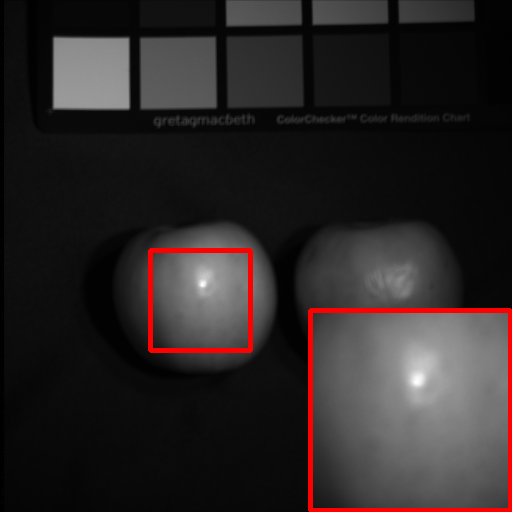} \\
		\quad
		\includegraphics[width=0.99999\columnwidth]{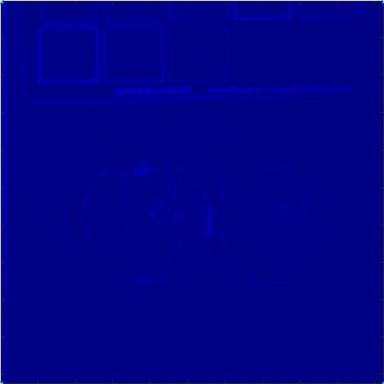}
		\subcaption{SRFN}
	\end{minipage}
	\begin{minipage}[t]{0.12\linewidth}
		\centering
		\includegraphics[width=0.99999\columnwidth]{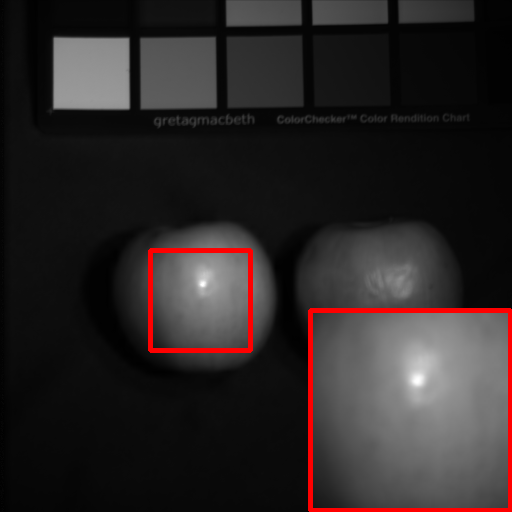} \\
		\quad
		\includegraphics[width=0.99999\columnwidth]{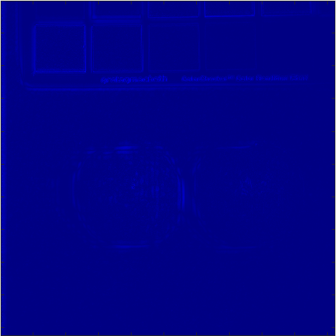}
		\subcaption{SRFN/GT}
	\end{minipage}
	\caption{Qualitative results of the CAVE dataset at band 31. Top row: reconstructed images. Bottom row: reconstruction errors -- light color indicates less error, dark color indicates larger error.}
	\label{fig:cave}
	\centering
\end{figure*}

We summarized the average quantitative results from 12 testing images in Table~\ref{tab:cave}. From the table, we can observe that the proposed method can obtain better reconstruction results against blind HIF method in terms of all four metrics. Further more, SRFN achieve better results than NSSR, which is a classical non-blind method. The non-blind version SRFN/GT achieve best performance. To visualize the results, we show the reconstructed samples of CAVE in Fig~\ref{fig:cave}. The difference images show that the proposed approach is able to better reserve both the spectral and spatial information. The SRFN obtains similar visual quality as the non-blind version, indicating that our method is able to accurately learn the observation model.

\begin{small}
\begin{table}[!t]
	\centering
	\setlength{\tabcolsep}{1mm}{
	\begin{tabular}{lccccc}
		\hline
		Method & Blind & PSNR$\uparrow$ & SSIM$\uparrow$ & SAM$\downarrow$ & ERGAS$\downarrow$ \\
		\hline
		HySure & \checkmark & 37.35  & 0.945 & 9.84 & 2.010 \\
		CuCaNet & \checkmark & 38.36  & 0.974 & 6.25 & 1.660 \\
		SRFN & \checkmark & 45.73  & 0.990  & 3.960 & 0.700\\
		NSSR & $\times$ & 43.82 & 0.987 & 4.070 & 0.840\\
		SRFN/GT & $\times$ & 45.97 & 0.991 & 3.288 & 0.590 \\
		\hline
	\end{tabular}}
	\caption{Quantitative comparison with SOTA unsupervised HIF methods on CAVE Dataset.}
	\label{tab:cave}
\end{table}
\end{small}

\begin{table}[h]
	\centering
	\begin{tabular}{lllll}
		\hline
		Method  & PSNR$\uparrow$ & SSIM$\uparrow$ & SAM$\downarrow$ & ERGAS$\downarrow$ \\
		\hline
		HySure  & 35.46  & 0.968 & 4.068 & 1.274 \\
		CuCaNet & 40.14 & {\color{red}{0.985}} & 2.943 & 0.831 \\
		SRFN   & {\color{red}{41.67}}  & 0.984  & {\color{red}{2.30}} & {\color{red}{0.820}}\\
		SRFN/GT   & 42.87  & 0.988  & 2.140 & 0.626\\
		\hline
	\end{tabular}
	\caption{Quantitative comparison with HySure and CuCaNet on Pavia Dataset. The best results are highlighted in red colors}
	\label{tab:qavia}
\end{table}

\begin{figure*}[!t]
	\centering
	\begin{minipage}[t]{0.13\linewidth}
		\centering
		\includegraphics[width=0.99999\columnwidth]{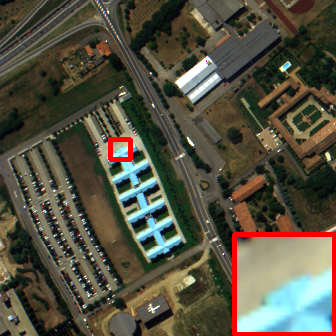}
		\subcaption{GT}
	\end{minipage}%
	\quad
	\begin{minipage}[t]{0.13\linewidth}
		\centering
		\includegraphics[width=0.99999\columnwidth]{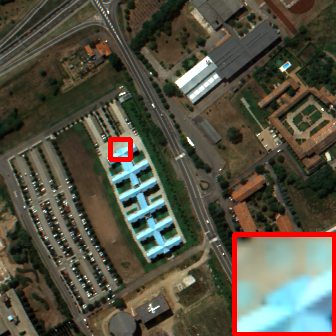} \\
		\quad
		\includegraphics[width=0.99999\columnwidth]{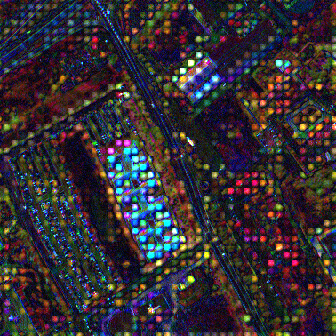}
		\subcaption{HySure}
	\end{minipage}
	\begin{minipage}[t]{0.13\linewidth}
		\centering
		\includegraphics[width=0.99999\columnwidth]{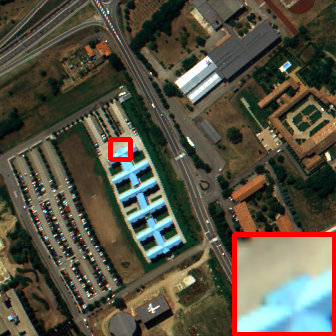} \\
		\quad
		\includegraphics[width=0.99999\columnwidth]{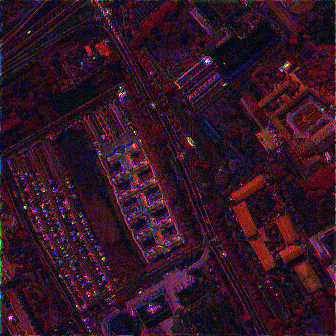}
		\subcaption{CuCaNet}
	\end{minipage}
	\begin{minipage}[t]{0.13\linewidth}
		\centering			
		\includegraphics[width=0.99999\columnwidth]{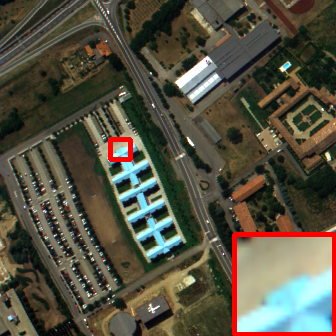} \\
		\quad
		\includegraphics[width=0.99999\columnwidth]{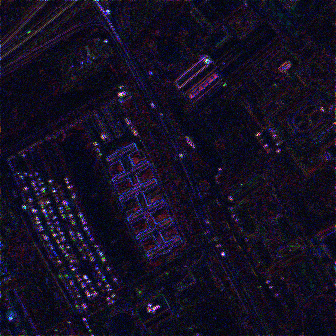}
		\subcaption{SRFN}
	\end{minipage}
	\begin{minipage}[t]{0.13\linewidth}
		\centering
		\includegraphics[width=0.99999\columnwidth]{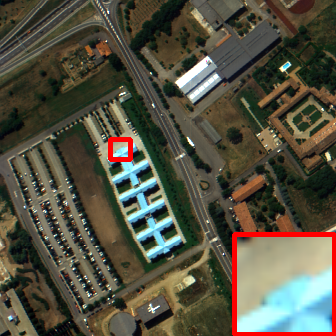} \\
		\quad
		\includegraphics[width=0.99999\columnwidth]{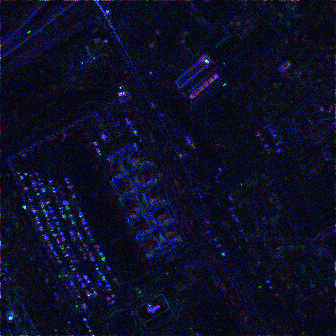}
		\subcaption{SRFN/GT}
	\end{minipage}
	\caption{Qualitative results of the Pavia dataset. The false-color image with bands 52-27-1 as R-G-B channels is displayed. Top row: reconstructed images. Bottom row: reconstruction errors. For clarity, we magnified the reconstruction error by 20 times}
	\label{fig:pavia}
	\centering
\end{figure*}

\begin{figure}[!t]
	\centering
	\begin{minipage}[t]{0.42\linewidth}
		\centering
		\includegraphics[width=1\columnwidth]{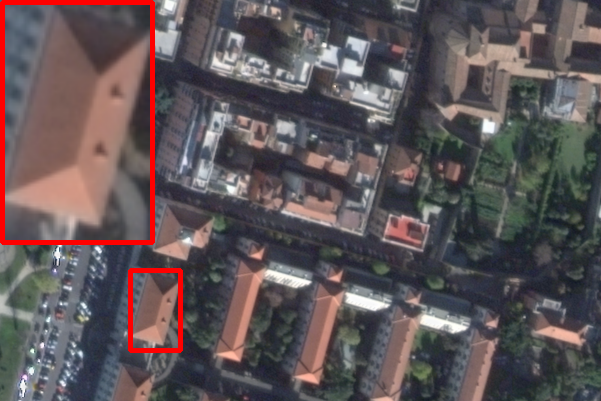}
		\subcaption{RGB}
	\end{minipage}
	\begin{minipage}[t]{0.42\linewidth}
		\centering
		\includegraphics[width=1\columnwidth]{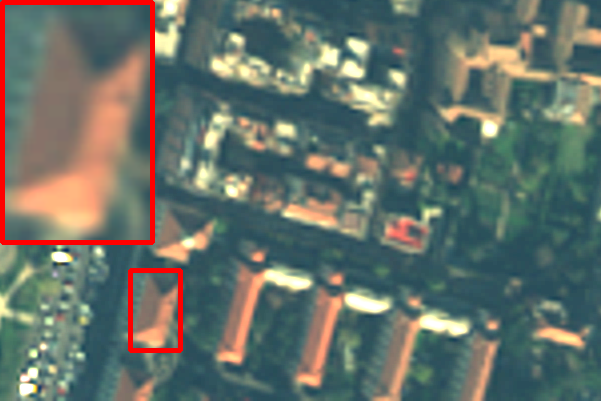}
		\subcaption{LR-MSI}
	\end{minipage}
	\begin{minipage}[t]{0.42\linewidth}
		\centering
		\includegraphics[width=1\columnwidth]{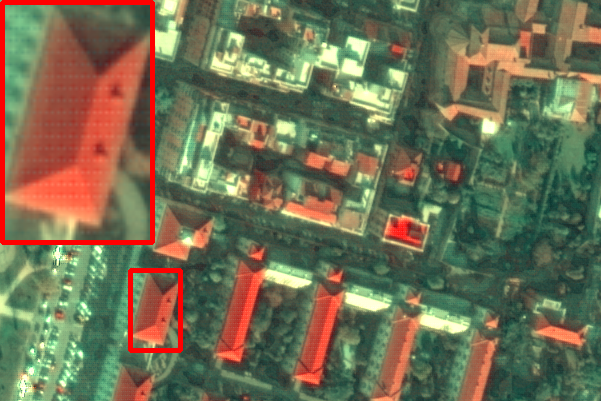}
		\subcaption{HySure}
	\end{minipage}
	\begin{minipage}[t]{0.42\linewidth}
		\centering
		\includegraphics[width=1\columnwidth]{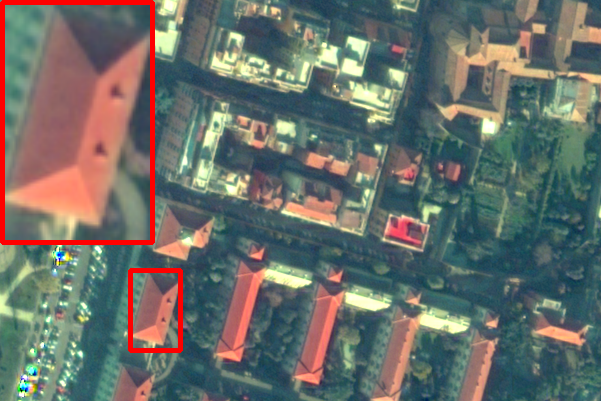}
		\subcaption{Ours}
	\end{minipage}
	\caption{(a) and (b) are the real RGB and LR-MSI image acquired by World View-2. (c)-(d) The reconstructed HR-MSI. The false-color image with bands 5-3-2 as R-G-B channels is demonstrated.}
	\label{fig:wv2}
	\centering
\end{figure}

\subsection{Experiments on Synthetic Remotely Sensed Dataset}
The synthetic indoor data contains only 31 spectral bands, while the remotely sensed data often contain hundred of spectral bands. To test the generality of our method, we carry out more experiments on synthetic remotely sensed data. We only compare our SRFN to the most related work, namely HySure and CuCaNet. The SRFN/GT serves as an up bound. Quantitative evaluation results of the Pavia University dataset are provided in Table~\ref{tab:qavia}. As shown in Table~\ref{tab:qavia}, SRFN constantly outperforms HySure and CuCaNet. In summary, SRFN delivers noticeable PSNR, SAM and ERGAS improvements on HySure and CuCaNet. The qualitative comparison is also illustrated in Fig.~\ref{fig:pavia}. HySure suffers from both spatial and spectral distortion, while CuCaNet achieves satisfactory results, difference image indicating that SRFN can still outperforms CuCaNet.

\subsection{Experiments on Real Remotely Sensed Dataset}
Besides the experiments on synthetic dataset, we further extend the experiments to real remotely sensed dataset to test the generality of our method. The experiments are carried on WV2 dataset. Comparison is made only with HySure since we could not achieve satisfactory results with the code provided by~\cite{cucanet}. The visual results are display in Fig.~\ref{fig:wv2}. HySure suffers from severe grid structure distortion and spectral distortion. In contrast, the proposed SRFN reconstructs a sharper and clearer quality. Our reconstructed image produces slightly spectral distortion, showing that further improvement could be made.

\section{Conclusion}

In this work, we proposed a self-regression learning method for blind HIF without label. The proposed SRFN is able to reconstruct a HR-HSI given a single pair of LR-HSI and HR-MSI in an unsupervised manner without knowing the observation model. By making part of the proposed network invertible, we are able to better preserve spectral and spatial information. Since the problem is highly under constraint, we proposed a local consistency loss to constrain the SRFN inspired by domain knowledge. We also introduce a spectral  reconstruction loss to improve spectral reconstruction accuracy. We have demonstrated the superior of the proposed SRFN in HIF. Besides, our network design is universal. With small modification, the proposed method can potentially be applied to other image processing problems such as blind image super-resolution or blind image deconvolution.

\bibliographystyle{IEEEtran}
\bibliography{egbib}

\end{document}